\documentclass[journal,transmag]{IEEEtran}
\usepackage{cite}

\ifCLASSINFOpdf
\usepackage[pdftex]{graphicx}
\usepackage{subfigure}
\else
\fi
\usepackage{amsmath}
\usepackage{amsfonts}
\usepackage{color}
\usepackage{xcolor}
\usepackage{multirow}
\usepackage{longtable}
\newcommand{\tabincell}[2]{\begin{tabular}{@{}#1@{}}#2\end{tabular}}
\usepackage {footnote} 
\makesavenoteenv {tabular}
\usepackage{threeparttable}
\usepackage{amsmath}
\usepackage{booktabs}
\usepackage{threeparttable}
\begin{document}
\title{Air-Writing Translater: A Novel Unsupervised Domain Adaptation Method for Inertia-Trajectory Translation of In-air Handwriting}


\author{Songbin Xu, Yang Xue, Xin Zhang, Lianwen Jin}

\markboth{Journal of xxx Class Files,~Vol.~1, No.~1, June~2019}%
{Shell \MakeLowercase{\textit{et al.}}: Bare Demo of IEEEtran.cls for IEEE Transactions on Magnetics Journals}

\IEEEtitleabstractindextext{%
\begin{abstract}
As a new way of human-computer interaction, inertial sensor based in-air handwriting can provide a natural and unconstrained interaction to express more complex and richer information in 3D space. However, most of the existing in-air handwriting work is mainly focused on handwritten character recognition, which makes these work suffer from poor readability of inertial signal and lack of labeled samples. To address these two problems, we use unsupervised domain adaptation method to reconstruct the trajectory of inertial signal and generate inertial samples using online handwritten trajectories. In this paper, we propose an Air-Writing Translater model to learn the bi-directional translation between trajectory domain and inertial domain in the absence of paired inertial and trajectory samples. 
Through semantic-level adversarial training and latent classification loss, the proposed model learns to extract domain-invariant content between inertial signal and trajectory, while preserving semantic consistency during the translation across the two domains. 
We carefully design the architecture, so that the proposed framework can accept inputs of arbitrary length and translate between different sampling rates. 
We also conduct experiments on two public datasets: 6DMG (in-air handwriting dataset) and CT (handwritten trajectory dataset), the results on the two datasets demonstrate that the proposed network successes in both Inertia-to Trajectory and Trajectory-to-Inertia translation tasks.
\end{abstract}

}

\maketitle

\IEEEdisplaynontitleabstractindextext

\IEEEpeerreviewmaketitle

\section{Introduction}

\IEEEPARstart{I}{n-air} handwriting refers to a novel way of human-computer interaction (HCI), which freely writes meaningful characters in 3D space and then converts them into user-to-computer commands. Compared with general motion gestures, in-air handwriting is more complicated and provides more abundant expressions. As modern MEMS(Micro-Electro-Mechanical System) inertial sensors become smaller and more energy efficient, they have been universally employed in portable and wearable devices such as smartphones and wristbands. Unlike optical devices, inertial sensors do not suffer from illumination interference and obstruction. Therefore, inertial sensor based in-air handwriting has widely attracted researchers' attention\cite{amma2012airwriting, lefebvre2015inertial, amma2014airwriting, murata2014hand}.

\begin{figure}[ht]
	\centering
	\includegraphics[width=0.48\textwidth]{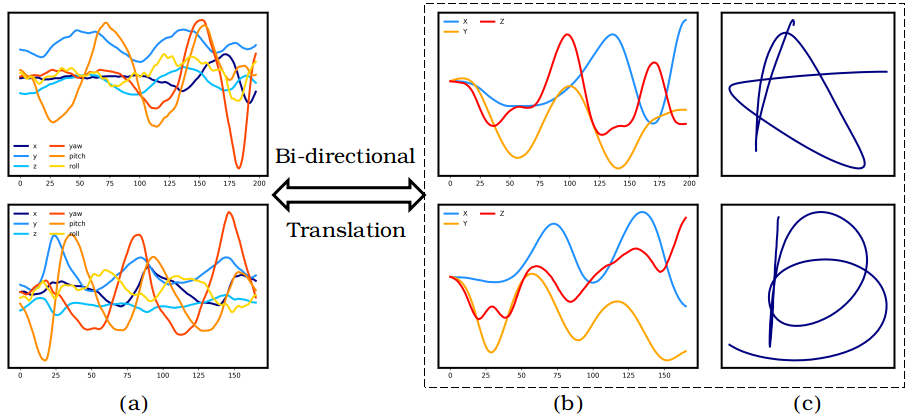}
	\caption{Illustration of inertial and trajectory signals. (a) The six-dimensional inertial data of in-air handwritten character, (b) The three-dimensional dynamic trajectory data of handwritten character, (c) Drawing the trajectory of x-y coordinates. The purpose of this paper is to achieve real and reliable translation between them.}
	\label{figure1}
\end{figure}

Most of the existing work is mainly focused on in-air handwriting recognition (IAHR)\cite{xu2017long,xu2016air,yana2018fusion, tsai2017reverse}. But in the research of IAHR, 

there are usually two problems. Firstly, the inertial signal is full of abstractness and lack of readability, because it is a series of temporal sequences representing motion shifting, as illustrated in Fig.\ref{figure1}(a). This characteristic makes it impossible to give an intuitive representation of the handwritten contents like an image. Furthermore, it may bring difficulty in sample labelling, character-level segmentation, or algorithms that benefit from data observation. Previous studies\cite{kaur2016mems, lee1993multiposition, wu2005strapdown} have attempt to reconstruct the trajectory of inertial signals using INS theory. However, integration errors accumulate and cause unavoidable drifts during trajectory reconstruction. In this paper, using the unsupervised domain adaptation method, we consider the trajectory reconstruction task as a translation task from an inertial signal to a trajectory signal. We define this task as \textbf{\emph{Inertia-to-Trajectory translation}}. Secondly, there aren't many labeled inertial in-air handwritten datasets. One of the reason may be that the acquisition cost of the labeled inertial samples is relatively high. Because during the data collection process, an extra optical tracking equipment is needed to record the handwritten trajectory for labelling. However, unlike inertial data, there are many traditional dynamic handwritten trajectory (a sequence of coordinates) datasets available. If inertial data can be generated using traditional trajectory data, the problem of insufficient inertial dataset can be solved. In this paper, we treat this generation task as \textbf{\emph{Trajectory-to-Inertia translation}}, which means a translation from a trajectory signal to an inertial signal.

Unlike image-to-image translation, in Inertia-Trajectory translation task (including Inertia-to-Trajectory and Trajectory-to-Inertia translations), inertial data and trajectory data are time series. On the one hand, the time series has variable length. How to handle sequences of arbitrary length as input is a problem. On the other hand, time series from different datasets are sampled at different sampling rates, which results in different duration distributions. In addition, in image-to-image translation, researchers mainly conduct sample-level adversarial training which requires proir knowledge that the pixel-level structure is roughly the same between two images. However, inertial data and trajectory data hardly share any pixel-level structure, which means that the existing image-to-image tranlation methods are not well suited for Inertia-Trajectory translation.

In this work, we propose a novel domain adaptation model named the \textbf{\emph{Air-Writing Translater}}, to address unsupervised Inertia-Trajectory translation (Bidirectional translation between Inertia and Trajectory) for character-level in-air handwriting. The proposed model mainly includes the following characteristics:

\begin{itemize}
\item The Air-Writing Translater is trained in an unsupervised way, in no need of pair-wise in-air handwriting datasets, since paired supervision is a strict constraint for practical applications. Inertial and trajectory samples in the same batch are not required to be paired, nor of the same writing style or character class.

\item The Air-Writing Translater combines adversarial training and classification in the feature level (semantic space) rather than the sample level to guide the semantic consistency. For Inertia-to-Trajectory translation, it can translate inertial signal into real and human-readable handwritten trajectory without modifying its semantic style. In the meanwhile, the translated trajectories stay the same writing style to the target domain, which is like handwriting style transfer. For Trajectory-to-Inertia translation, it can convert dynamic trajectory into real and diverse inertial in-air handwriting data.

\item The Air-Writing Translater can obtain a fixed-length semantic feature through Conv and GRU, regardless of the length of the input sequences, which makes it easier to conduct semantic consistency. In addition, this unique design benefits our model in another way: the model can convert between inertial data and trajectory data collected at different sampling rates. 

\end{itemize}

In particular, we design a two-stream ConvNet\cite{simonyan2014two} that combines the translated samples from the target domain with the source domain samples, to significantly improve the classification performance.

To the best of our knowledge, this is the first work on addressing unsupervised Inertia-Trajectory translation as a domain adaptation task. The experimental results on two public datasets CharacterTrajectories\cite{williams2006extracting} and 6DMG\cite{chen2015air} demonstrate that the Air-Writing Translater can effectively achieve reliable translation between inertial signal and handwritten trajectory.

\section{Related Work}

In recent years, literature about inertial sensor based in-air handwriting pays more attention on realizing accurate classification\cite{chen2015air, lefebvre2015inertial}, rather than discussing the inertial signal itself. The visualization task for inertial signal is common in Inertial Navigation System\cite{kaur2016mems, lee1993multiposition, wu2005strapdown} and Inertial Tracking\cite{filippeschi2017survey}, serving as a trajectory reconstruction task, which aims to develop inertial sensory data to estimate real-time position and orientation. The major disadvantage is the poor accuracy\cite{chen2018transferring}, due to the accumulated noise and drift error in inertial measurement units. Researchers solve these issues with filters and compensation methods, such as Kalman filters\cite{huang2017kalman}\cite{nguyen2017developing} and Zero Velocity Compensation (ZVC)\cite{wagstaff2018lstm}\cite{wang2017smartphone}. Contrary to previous work, we consider inertial visualization from a novel perspective. Specifically, we represent inertial sensory signal with handwritten trajectory of the same semantic style, rather than directly recovering the real trajectory. Without sufficient research on Inertia-Trajectory translation as reference, we draw inspiration from literature in image area about domain adaptation and unsupervised image-to-image translation.

\subsection{Style Transfer}
In image area, neural style transfer attempts to modify the style (textures, details) of an image, while keeping its structural content (layouts, contours) unchanged. Earlier work focus on example-guided style transfer\cite{gatys2016image}\cite{li2017universal}, conditioned on both input image and an example from the target domain. Some recent work also allow transfer between very dissimilar domains\cite{li2016combining}\cite{johnson2016perceptual}. Basically, the inputs and outputs of a style transfer system share the same pixel layout. However, inertial signal and handwritten trajectory are totally different in physical sense, the former is a relative measurement of motion shifting, while the latter a positional record of the movement itself. There are thus hardly any similar pixel layout features shared by the two domains. Therefore, pixel-level style transfer is not suitable for Inertia-Trajectory translation.

\subsection{Unsupervised Image-to-Image translation}
Our work is close to unsupervised image-to-image translation in terms of partial objectives. The CycleGAN\cite{zhu2017unpaired} separately applies an image converter for each domain, and learns image-to-image translation through a cycle-consistency loss and adversarial training. This pipeline constrains nothing on the relationship across domains, but relies on pixel-level layouts learned in self-supervision. It may lead to unexpected corresponding if applied to Inertia-Trajectory translation, where shared pixel structures can hardly be found. Some researchers make constraints to guide the transfer, so are we. The DTN\cite{taigman2016unsupervised} realizes face-to-cartoon translation with a single auto-encoder, where the encoder is pretrained as a feature extractor for both domains. Such a restriction of reusing encoder is infeasible in our scene, because inertial signal and trajectory are sequences of different dimension and arbitrary length. The UNIT\cite{liu2017unsupervised} is a coupled VAE-GAN, it translates images through a domain-sharing semantic space. The major difference is, UNIT makes a weight-sharing constraint in network architecture, while we realize semantic consistency by optimizing explicit loss functions. The XGAN\cite{royer2017xgan} applies a domain classifier on the top of embedding bottleneck and trains via a gradient reversal layer, while our work uses a latent discriminator which is trained jointly with domain encoders like GAN. Note that the researches mentioned above contain sample-level adversarial training after translated images, while ours at feature-level.

\subsection{Unsupervised Domain adaptation}
To some extents, domain adaptation is like semantic-level style transfer, because it extracts domain-invariant features and learns a feature-level mapping between domains. Ganin et al.\cite{ganin2014unsupervised} introduces Domain-Adversarial Neural Network (DANN), aiming at a classifier discriminative in both source and target domain. By introducing a domain classifier, DANN makes the features extracted from both domains be similar. As advanced versions of DANN, \cite{long2015learning} and \cite{tzeng2014deep} replace the domain classification loss with Maximum Mean Discrepancy. The above work concentrates on pixel translation, and most of them target classification but not domain adaptation itself. 

Finally, the MotionTransformer\cite{chen2018transferring} is very close to this work in terms of application. The MotionTransformer addresses a scene of Inertial Tracking, which utilizes linear and angular acceleration to recover the walked trajectory. The framework solves it as a domain adaptation task without any paired training data. Unlike our work however, MotionTransformer only considers unidirectional translation, and it introduces sample-level adversarial training but not feature-level.

\section{Inertia and Trajectory}
In this work, we focus on character-level in-air handwriting, where each sample is sequential and corresponds to an isolated handwritten character. In this section, we intend to provide some concepts about Inertia-Trajectory translation.

\subsection{Bi-Directional Translation}
For character-level in-air handwriting, the Inertia-to-Trajectory translation refers to a procedure that translates inertial signal into handwritten trajectory. And the Trajectory-to-Inertia translation is just reverse. The inertial signal is a six-dimensional sequence consisting of tri-axial acceleration and tri-axial angular velocity. While the trajectory is a sequence of three dimensional coordinates that can be drawn one by one on a plane to make it human-readable.

\subsection{Domain Concept}
The concept of domain is common in unsupervised image-to-image translation\cite{liu2017unsupervised}\cite{royer2017xgan}, and domains usually refer to different attributes or aspects of the same object. In this work, we consider inertial signal and handwritten trajectory as samples from two different domains, because they are essentially different representations of in-air handwriting. The translation task attempts to convert a sample from one domain to another.

\begin{figure*}
\centering
\subfigure[Architecture of Air-Writing Translater]{
\begin{minipage}[b]{0.54\textwidth}
\includegraphics[width=\textwidth]{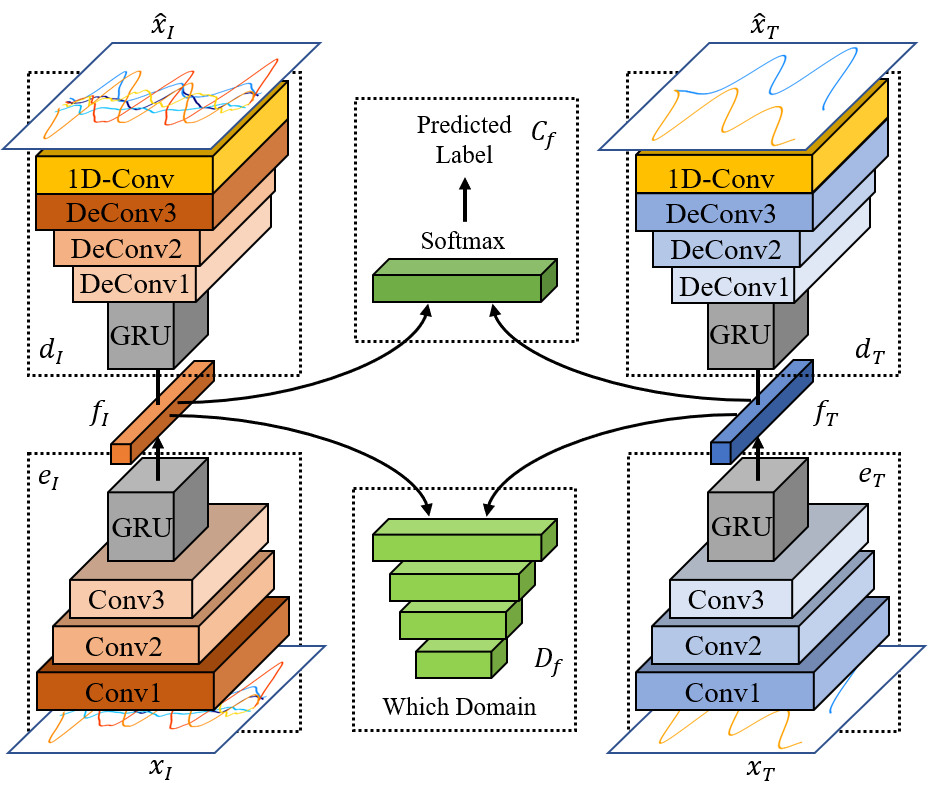}
\end{minipage}
}
\subfigure[Illustration of Inertia-to-Trajectory translation]{
\begin{minipage}[b]{0.4\textwidth}
\includegraphics[width=\textwidth]{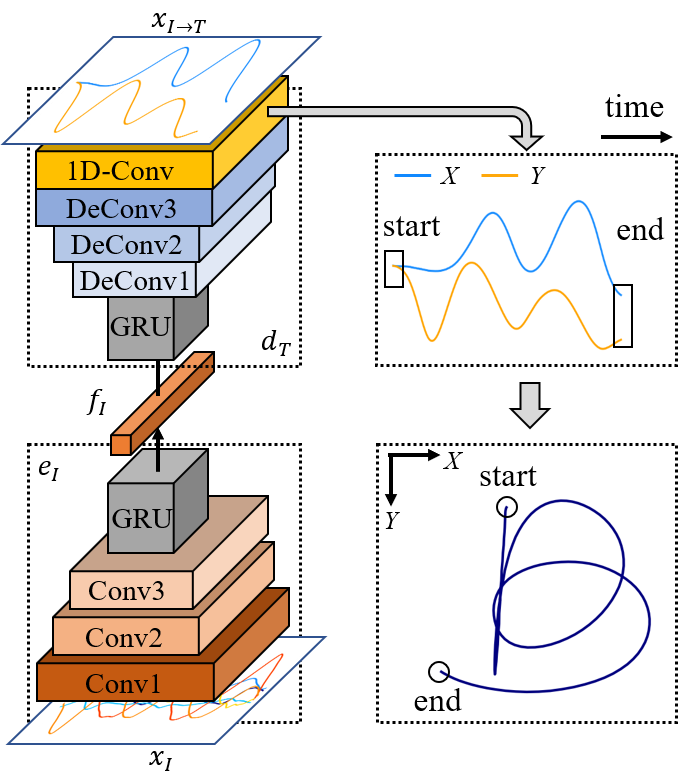}
\end{minipage}
}
\caption{(a) Architecture of Air-Writing translater. (b) Illustration of Inertia-to-Trajectory translation. (a): $x_I$ and $x_T$ denote samples from domain I and T. $e_I$ and $d_I$ denote encoder and decoder of domain I. $e_T$ and $d_T$ denote encoder and decoder of domain T. $f_I$ and $f_T$ denote the fixed-length semantic content extracted by $e_I$ and $e_T$, respectively. $\hat{x}_I$ and $\hat{x}_T$ represent samples reconstructed by auto-encoder means samples recovered by Auto-encoder $e_I$+$d_I$ and $e_T$+$d_T$, respectively. $C_f$ and $D_f$ represents represent a latent classifier and a latent discriminator, respectively, shared by domain $I$ and $T$. (b): $x_{I\rightarrow{T}}$ represents a reconstructed trajectory from inertial domain by using $e_I$ and $d_T$.}
\label{figure2}
\end{figure*}

\subsection{Unpaired Sample and Unsupervision}
In general, the data collection of inertial sensor based in-air handwriting provides only inertial signal samples. Some researchers introduce an extra optical tracking equipment to simultaneously record the real-time hand-written trajectory, such as M.Chen et.al\cite{chen2015air}. Therefore, each inertial sample is accompanied by a trajectory sample, which we refer to as \textbf{paired sample}. Any other unsynchronized recorded trajectory cannot be seen a paired trajectory, even if it is of the same character class or written by the same person.

If we introduce the concept "unsupervised" from image-to-image translation, we can define that the training with unpaired inertial sample and trajectory sample is unsupervised. In this paper, we focus on unsupervised Inertial-Trajectory translation.

\subsection{Semantic Content}
The semantic content of an image is what it describes, we can summarize it by directly observing, such as an object with some attributes or an event. However, it's difficult to define semantic content for in-air handwriting. Because it's hardly possible to conclude anything by observing a segment of inertial signal, most of the time we see nothing but noises. Therefore, we consider semantic content as the core topic that expresses what the sample exactly is. For character-level in-air writing, we define its semantic content as the character class. We introduce a semantic space shared by samples from both domains, to ensure that the semantic content is domain-invariant.

\section{METHODOLOGY}

\subsection{Motivations}
Although GANs have achieved outstanding performance in unsupervised image-to-image translation, they are still limited by the lack of capability of handling sequential data. Previous methodologies which have been proved effective in image area cannot be directly adapted to in-air handwriting, there are still several challenges to be solved.

\textbf{\emph{Lack of paired training data}}. In supervised settings, it's feasible to train a sequence-to-sequence model with sufficient paired data. For Inertia-to-Trajectory, such a model is end-to-end trainable as long as we take inertial signal as input and paired handwritten trajectory as target sequence. However, a pair-wise database requires synchronous data collection of motion trajectory, which brings extra costs on equipment and also burdens collectors. It would be better if we only need to access datasets of pure inertial signal and handwritten trajectory separately. Therefore, we prefer to train an unsupervised translation network. 

\textbf{\emph{Handling time series}}. Both inertial signal and trajectory (a sequence of coordinates) are time series. On one hand, in each domain, the sample has arbitrary length, which means samples of the same character class still last for different durations. Therefore, we prefer a tolerant framework which lays no constraints on data size. Besides, the duration distributions of inertial signal and trajectory may differ a lot if they are collected at very different sampling rates. There are two methods to solve this problem: (1) fixing the duration through interpolation or padding, (2) designing a flexible network which adaptively decide the translated duration. Interpolation is not appropriate for inertial signal because its amplitude is strongly relevant to duration, and padding is not flexible for practical applications. Therefore, we prefer to rely on the model's flexibility.

\textbf{\emph{Difficulty in sample-level adversarial training}}. Previous works mainly introduce adversarial training between translated images and target domain images\cite{taigman2016unsupervised}-\cite{royer2017xgan}. By default, the input image and translated image somehow share the same pixel-level structure. These styles are not transferred but more like preserved. For example, in horse-to-zebra\cite{zhu2017unpaired} the input and output share the entire layout, background and object outlines, only differ in texture and color of the horse. However, the inertial signal and trajectory are totally different in physical sense, data size and details. They hardly share any pixel-level styles, except for the semantic content. To this end, we propose to simply ignore the specific styles existing in each domain, but instead transfer semantic content. In other words, we prefer a domain adaptation model which conducts semantic-level adversarial training.

\subsection{Proposed Model: Air-Writing Translater}
Let $I$ and $T$ denote the inertial domain and trajectory domain, respectively. The architecture of our proposed model is shown in Fig.\ref{figure2}(a). The proposed Air-Writing Translater has a symmetric structure, and consists of three modules, including a pair of auto-encoders ($e_I$ and $d_I$ denote encoder and decoder of the inertial domain, $e_T$ and $d_T$ denote encoder and decoder of the for trajectory domain), a latent classifier $C_f$ (shared by domain $I$ and $T$) and a latent discriminator $D_f$ (shared by domain $I$ and $T$).

Given an inertial sample $x_I$ with a character label $y_I$, Air-Writing Translater maps $x_I$ into semantic representation $f_I$, which is shared by 3 parts. Firstly, the inertial decoder $d_I$ uses it to produce a reconstructed inertial sequence $\hat{x}_I=d_I(f_I)$. Secondly, the latent classifier uses it to try to recover the label $\hat{y}_I=C_f(f_I)$. Finally, the latent discriminator uses it to calculate $D_f(f_I)$ to indicate which domain $f_I$ comes from. Since the proposed model is symmetric, the process of domain T is exactly the same as domain I. 

We take a two-steps strategy to interconnect two different domains instead of converting directly between them. The Inertia-to-Trajectory translation is illustrated in Fig.\ref{figure2}(b). Firstly, the inertial encoder $e_I$ extracts semantic content $f_I$ from inertial signal $x_I$ , where the semantic latent space serves like a transfer station between two domains. Secondly, the decoder $d_T$ reconstructs a sequential trajectory $x_{I\rightarrow T}=d_T(f_I)$ from the semantic latent representation.

Next, we intend to explain why we design the Air-Writing Translater like this.

\textbf{\emph{Feature Extraction}}. Our first target is to extract important styles from inertial signal and trajectory. We introduce an auto-encoder for each domain, which is famous as unsupervised feature compression model. Take inertial domain for example, $e_I$ attempts to extract necessary features $f_I$ from $x_I$, so that the decoder $d_I$ can recover $x_I$ from $f_I$. The two auto-encoders are trained by optimizing a sample-level reconstruction loss $L_{rec}$. Addressing sequential data, the loss is an element-wise $L1$ norm between the original and the reconstructed sequence. With the $L1$ objective, losses along timesteps tend to be sparse, leading to a smoother reconstruction. The $L_{rec}$ loss is separately calculated in each domain and then added up.
\begin{equation}
L_{rec}=\mathbb{E}\left\|x_I-d_I(e_I(x_I))\right\|_1+\mathbb{E}\left\|x_T-d_T(e_T(x_T))\right\|_1
\end{equation}

\textbf{\emph{Semantic Content Extraction}}. The second issue is to ensure that auto-encoder extracts semantic features. As we defined above, the semantic feature for in-air handwriting is the character class. Therefore, we introduce a latent classifier $C_f$, which accepts both $f_I$ and $f_T$ and attempts to predict the character label. By doing so, two encoders must compress character class information into the latent representations $f_I$ and $f_T$. This target is achieved by optimizing a latent classification loss $L_{cls}$, a sum of cross-entropy loss in each domain.
\begin{equation}
L_{cls}=-\mathbb{E}[y_I\log [C_f(e_I(x_I))]]-\mathbb{E}[y_T\log[C_f(e_T(x_T))]]
\end{equation}

\begin{figure}
\centering
\includegraphics[width=0.4\textwidth]{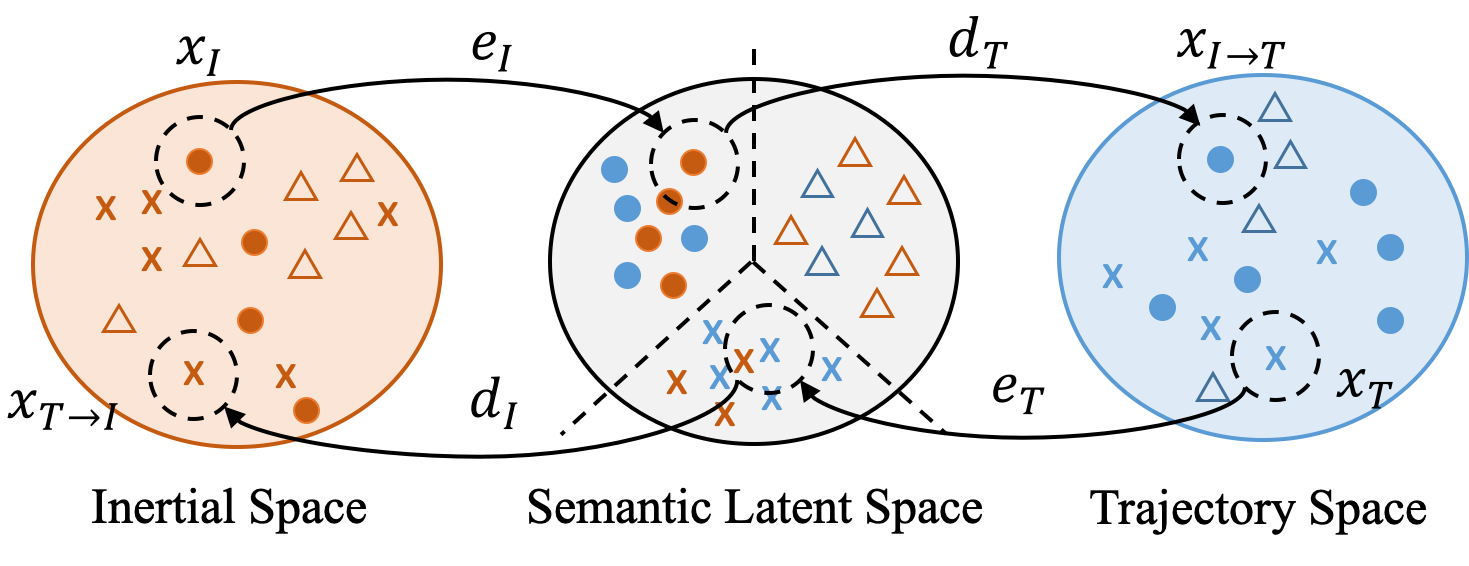}
\caption{Illustration of semantic content guidance. The latent classification loss $L_{cls}$ divides the latents according to semantic content, and the latent adversarial loss $L_{gan}$ brings latents of the same semantic content together.}
\label{figure3}
\end{figure}

\textbf{\emph{Semantic Consistency}}. We introduce a semantic-level adversarial training between latent $f_I$ and $f_T$ to guide semantic consistency. The inertial encoder $e_I$ acts as a generator that accepts inertial samples and provides semantic latents as generated samples, so does the trajectory encoder $e_T$. Following the ideas of GAN, we train $e_I$ and $e_T$ against the latent discriminator $D_f$. On one hand, $D_f$ aims to distinguish inertial domain latent $f_I$ from trajectory domain latent $f_T$, in order to determine which domain $f$ is from. On the other hand, the two encoders $e_I$ and $e_T$ attempt to confuse $D_f$ by generating similar $f_I$ and $f_T$ in probability distribution. Intuitively, adversarial training encourages each encoder to filter the information dependent to its domain in the latent representations and to preserve domain-invariant features. Through adversarial training, the latent representation in one domain becomes close to those latent representation in the other domain of the same character class. Therefore, the semantic consistency is achieved. Fig.\ref{figure3} shows how we guide semantic consistency. We alternatively train $e$ and $D_f$ to optimize a min-max objective.
\begin{equation}
\begin{split}
\min_{e}\max_{D_f}L_{gan}(e_I,D_f)=&\mathbb{E}_{x\sim p_I}[(D_f(e_I(x))-1)^2]+\\
&\mathbb{E}_{x\sim p_T}[D_f(e_T(x))^2]
\end{split}
\end{equation}

\textbf{Why is unsupervised}. In each iteration of batch-training for the Air-Writing Translater, we pass a batch of inertial samples into $e_I$  and a batch of trajectories into $e_T$. We don't require samples in two batches to be paired with each other. In fact, samples in the same order can be of different character types, writing styles and durations. The reason is obvious. The architecture design constrains weak relevance between inertial batch and trajectory batch, since they are mutually independent during the update of $L_{rec}$ and $L_{cls}$, except for $L_{gan}$. Furthermore, the nature of adversarial training determines that the proposed model learns to bring latent representations together from the perspective of probability distribution, rather than learning a one-to-one matching.

\begin{figure}
\centering
\begin{minipage}[b]{0.45\textwidth}
\includegraphics[width=\textwidth]{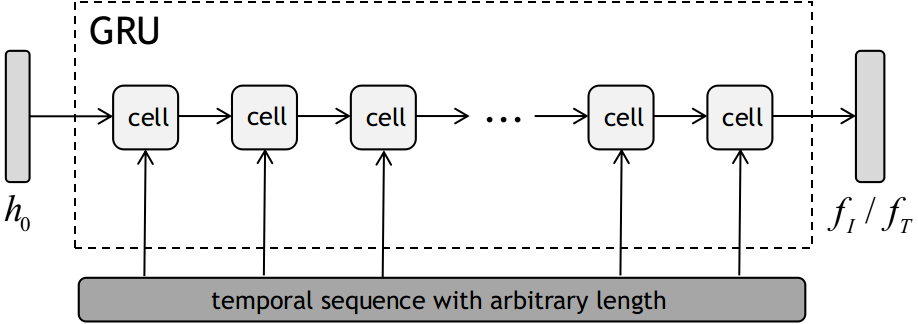}
\end{minipage}

\caption{Architecture of our recurrent blocks. $h_0$ means RNN's initial hidden state. $f_I$/$f_T$ represents latent representation with fixed length in both domains.}
\label{figure4}
\end{figure}

\textbf{How to handle time series and translate between different sampling rates}. There are two key issues to be solved: (1) mapping the time series into semantic latent vector of fixed-size, (2) generating the temporal sequence from fixed-size latent vector. For the first question, following the ideas of CRNN (B. Shi et.al, 2016), we set the encoder in the order of the convolution layers and then recurrent blocks (in this paper, we use GRU). After the convolution layers' processing, the temporal sequence still has variable length. Then, the GRU replaces the time series with the hidden states of the final time step, thereby realizing the conversion from the variable length semantic space to the fixed length semantic space. The architecture of GRU is shown in Fig.\ref{figure4}. For the second question, we assign recurrent blocks at the entrance of the decoders. In addition, through the unique design of the encoder and decoder, our model can naturally deal with samples from domains with different sampling rates. Decoder with recurrent blocks can generate samples that is close to the length of the source domain no matter if the length of the generated samples is reasonable in the target domain.

\subsection{Architecture Details}
\subsubsection{Auto-encoder}
We assign an auto-encoder for each domain, the weights are not shared but their major architecture is the same. The encoder consists of a down-sampling component with 3 convolutional layers and an embedding component with a single-layer GRU (Gated Recurrent Unit) network. The decoder is basically a mirror of the encoder, it firstly decodes the latent vector with GRU, and treats the emitted sequence as a temporal feature map, then up-samples it with 3 deconvolutional layers. We add one extra 1-D convolutional layer after the decoder, serving as a weighted filter which leads to smoother reconstruction and translation. The dimension of semantic latent vector is 64.

The latent vector is the last hidden state emitted by the GRU in encoder, and the decoder requires an assignment of output duration $t$ when it translates the vector. It copies the vector for $t$ times, and inputs the sequence into the GRU one by one. In auto-decoding, the duration $t$ is equal to the length of the last feature map in encoder. 

Note that we use raw inertial signal and trajectory data as inputs, which have a very large length-width ratio. Therefore, we use strip-shaped kernels in our network, which pays more attention to patterns along the time axis. In particular, we apply 1-D convolution except for the last convolutional layer in encoder and the first deconvolutional layer in decoder.

\subsubsection{Latent classifier and discriminator}
The latent discriminator is a feed-forward neural network, including 3 fully connected layers with leaky-ReLU activation and a linear layer. We assign 64 neurons on the output layer, so that the discriminator can make more reliable decision such as voting. The latent classifier is a single softmax layer, because extra hidden layers decrease the separability in latent space.

\section{Experiments}
In this section, we evaluate our Air-Writng Translater model on two public datasets. One is an in-air handwritten character dataset named \textbf{\emph{6DMG}}\cite{chen2015air}. The other is a traditional dynamic handwritten character dataset called CharacterTrajectory (\textbf{\emph{CT}})\cite{williams2006extracting}.

\subsection{Datasets}
\begin{figure}
\centering
\subfigure[]{
\begin{minipage}[b]{0.21\textwidth}
\includegraphics[width=\textwidth]{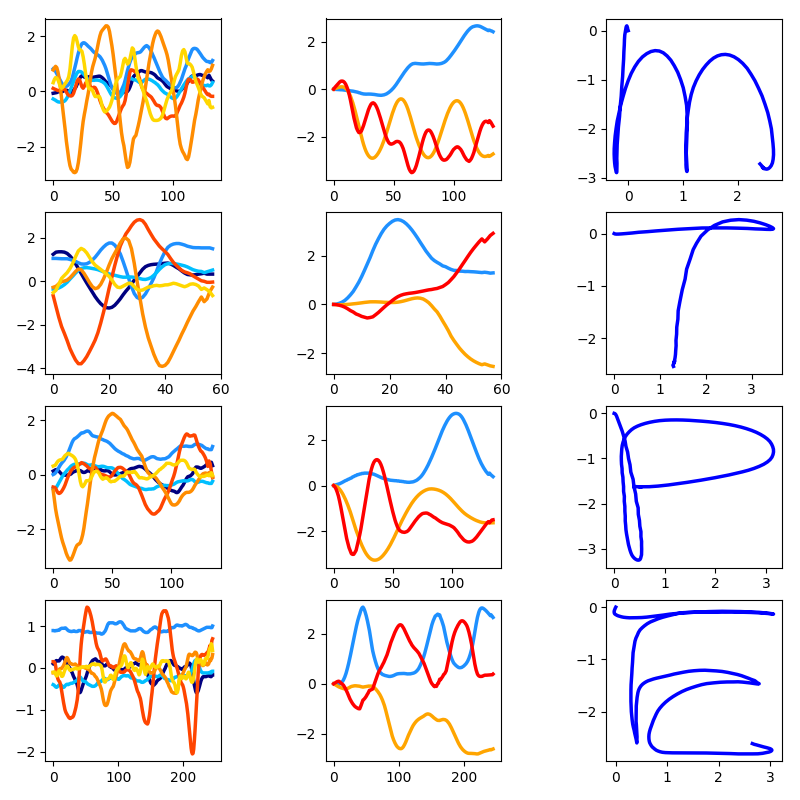}
\end{minipage}
}
\subfigure[]{
\begin{minipage}[b]{0.21\textwidth}
\includegraphics[width=\textwidth]{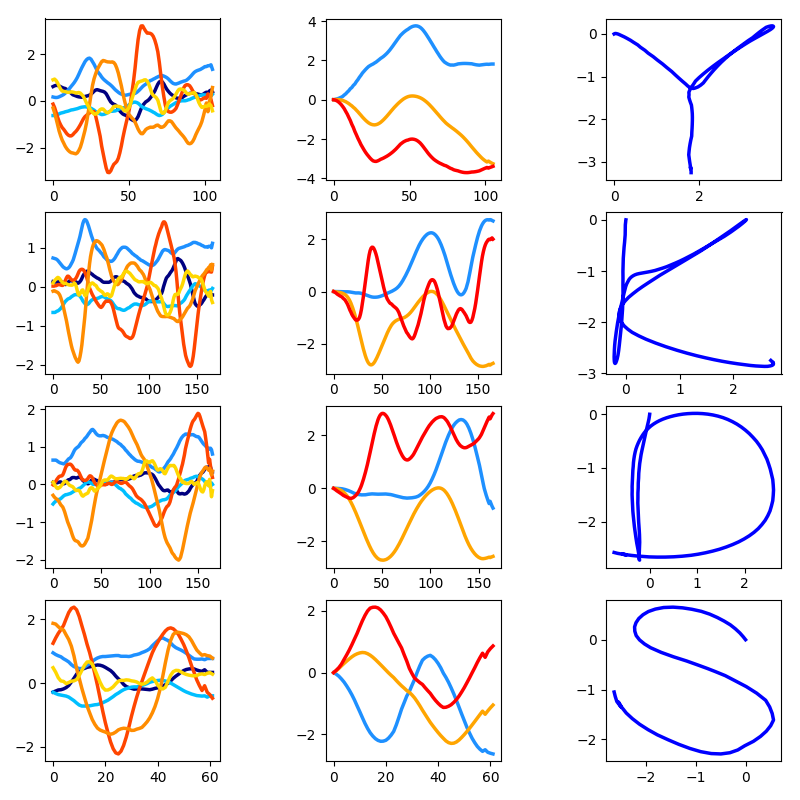}
\end{minipage}
}
\caption{(a) (b) are examples from 6DMG. 6DMG is a dataset of paired samples including inertial samples and trajectory samples. The fist column shows the waveform of the six-dimensional inertial signals. The Second colomn shows the waveform of the three-dimensional spatial trajectory data. In the last column, we plot the trajectory of the x-y coordinates in the plane Cartesian coordinate system.}
\label{figure5}
\end{figure}

\textbf{\emph{6DMG dataset}}: The 6DMG dataset is a collection of in-air handwritten characters. It contains 62 character classes, including 26 uppercase letters, 26 lowercase letters and 10 Arabic numerals. It contains 8,570 samples in total, including 600 samples of Arabic numerals, 6500 samples of uppercase letters and 1470 samples of lowercase letters. Each sample is a temporal sequence of an isolated in-air handwritten character, and the sequence length varies from each other. The 6DMG was collected under a hybrid framework, which recorded tri-axial acceleration and angular velocity using an inertial sensor, and captured trajectory of spatial coordinates with an optical tracking device. Therefore, the 6DMG contains both inertial domain signals and trajectory domain data. Fig.\ref{figure5} shows some examples from the 6DMG. We can see that the two domains look different in dimensions and trends. The six-dimensional inertial signal is not human-readable and it is difficult  to directly distinguish the character class by observing the waveform. However, the trajectory data is human-readable, and the xy coordinates of the trajectory are plotted in the last column of Fig.\ref{figure5} One of the purpose of this paper is to convert abstract inertial handwritten data into human-readable handwritten trajectory.

\begin{figure}
\centering
\subfigure[]{
\begin{minipage}[b]{0.14\textwidth}
\includegraphics[width=\textwidth]{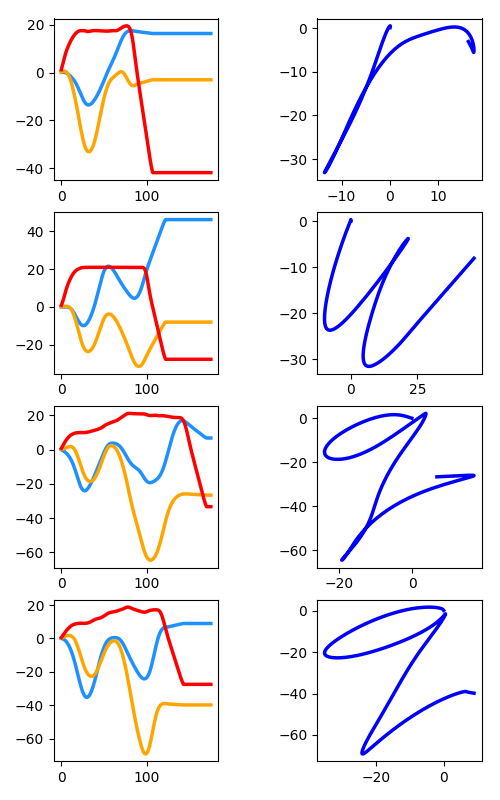}
\end{minipage}
}
\subfigure[]{
\begin{minipage}[b]{0.14\textwidth}
\includegraphics[width=\textwidth]{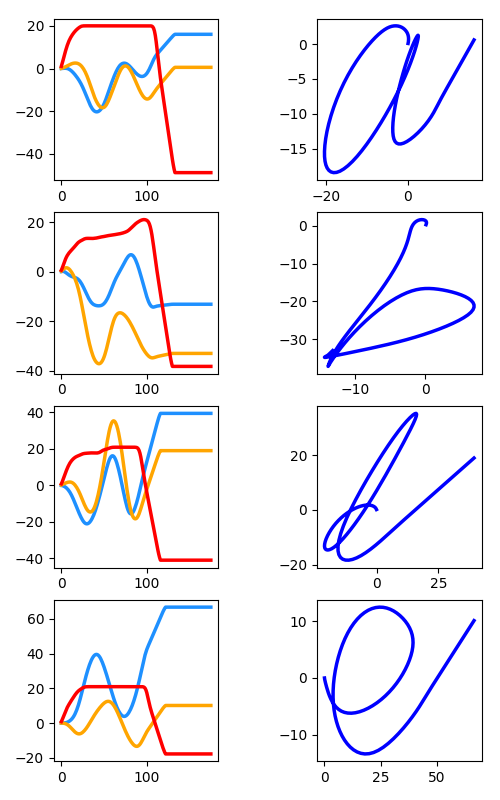}
\end{minipage}
}
\subfigure[]{
\begin{minipage}[b]{0.14\textwidth}
\includegraphics[width=\textwidth]{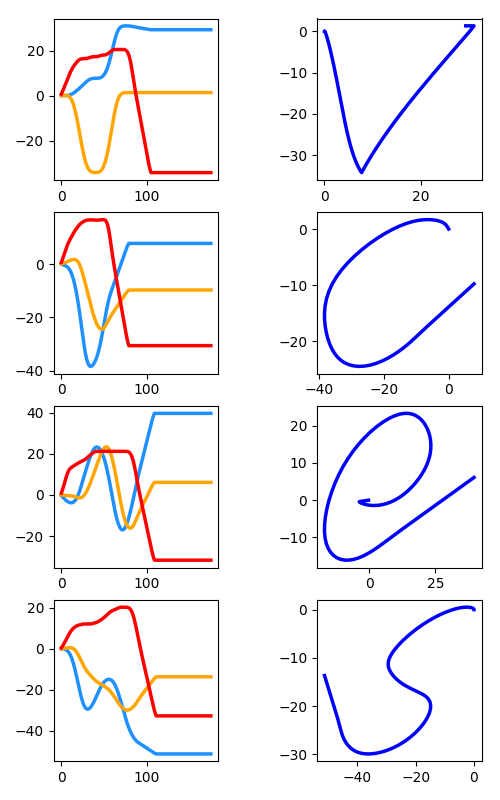}
\end{minipage}
}
\caption{(a)(b)(c) are examples from CT. CT only contains trajectory samples. The first column shows the waveform of the spatial trajectory data and the last column shows the trajectory of x-y coordinates plotted in the plane Cartesian coordinate system.}
\label{figure6}
\end{figure}

\textbf{\emph{CharacterTrajectory (CT) dataset}}: The CT dataset is a traditional handwritten trajectory dataset consisting of 20 classes of lowercase letters, with a total of 2,858 samples. The number of samples for each class is approximately equal. The CT was collected on a trackboard, and contains only dynamic trajectory samples. We show some trajectory examples randomly selected from the CT in Fig.\ref{figure6}. Similarly, the xy coordinates of the trajectory are drawn in the rightmost column.

In addition, the two datasets were collected at different sampling rates (60Hz in 6DMG and 200Hz in CT.)

\subsection{Datasets Pre-processing}
Addressing the characteristics of inertial and trajectory data, we performed different necessary data preprocessing. For inertial data, we perform a moving average filtering with a fixed window length of 5 to alleviate noise. For trajectory data, we subtract the coordinates of the first time step from each sequence so that all sequences start at zero origin.

\subsection{Experiment Setting}
\begin{figure}
\centering
\subfigure[Trajectory examples from 6DMG]{
\begin{minipage}[b]{0.48\textwidth}
\includegraphics[width=\textwidth]{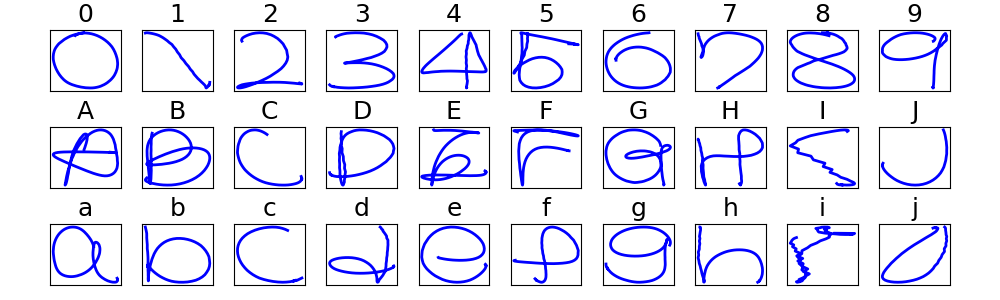}
\end{minipage}
}
\subfigure[Trajectory examples from CT]{
\begin{minipage}[b]{0.48\textwidth}
\includegraphics[width=\textwidth]{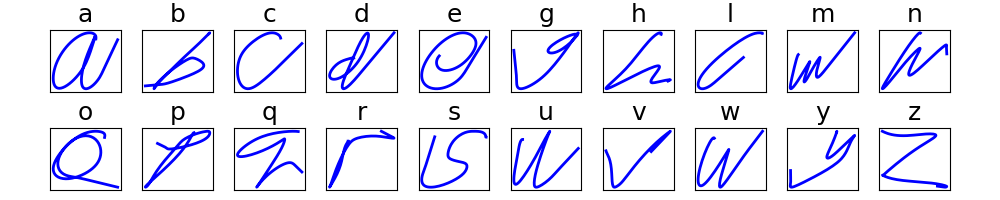}
\end{minipage}
}
\caption{Trajectories randomly sampled, presented in X-Y coordinate.}
\label{figure7}
\end{figure}

We evaluated our proposed Air-Writing tanslater under different experimental settings. According to different source and target domains, we divide the experimental settings into four groups, as shown in TABLE I. For the first and third sets of settings, data in different domains comes from the same dataset with the same handwriting style. To build an unsupervised circumstance, we first divide the paired examples into two sets of inertial and trajectory data, then shuffle them in a different order. By doing so, in each iteration the inertial and trajectory batches are not aligned, which forces our model to learn rather than remember the mapping. In the second and fourth sets of settings, we explored the feasibility of domain translation between different datasets with different handwriting styles. Fig.\ref{figure7} illustrate some of the character trajectory from 6DMG and CT. Compared to the 6DMG, the character trajectory of the CT is handwritten in a significantly different style. To remain the alignment of character class, we only pick out the samples of the 20 classes contained in CT from 6DMG. We take 80\% of the samples in each dataset for training. And the remaining 20\% is used to test the quality of the generated sample.

\renewcommand{\arraystretch}{1.0} 
\begin{table}

  \centering
  \caption{Experiment setting}
    \begin{tabular}{cccc}
    \toprule
    \tabincell{c}{Experiment\\setting} & \tabincell{c}{Source\\Domain} & \tabincell{c}{Target\\Domain} & Translation\cr
    \midrule
    1.Iner.-Tra.(6DMG) & \tabincell{c}{Inertial data\\from 6DMG} & \tabincell{c}{Trajectory data\\from 6DMG} & \multirow{4}{*}{\tabincell{c}{Inertia-\\to-\\Trajectory}}\cr\\
2.Iner.-Tra.(CT) & \tabincell{c}{Inertial data\\from 6DMG} & \tabincell{c}{Trajectory data\\from CT}\cr
\\
\hline
\\
3.Tra.(6DMG)-Iner. & \tabincell{c}{Trajetory data\\from 6DMG} & \tabincell{c}{Inertial data\\from 6DMG} & \multirow{4}{*}{\tabincell{c}{Trajectory-\\to-\\Inertia}}\cr\\
4.Tra.(CT)-Iner. & \tabincell{c}{Trajetory data\\from CT} & \tabincell{c}{Inertial data\\from 6DMG}\cr
    \bottomrule
    \end{tabular}
\end{table}


Although our training objective is to minimize total losses, we decompose them and train separately. Specifically, in one iteration, we successively optimize $L_{rec}$, $L_{cls}$ and $L_{gan}$, instead of updating the network jointly. During the optimization of each loss, only the relevant components are trained, while the rest are fixed. For instance, when optimizing $L_{cls}$, we only update weights in two domain encoders and the latent classifier. For adversarial balance, we update the latent discriminator 3 times in one iteration. Finally, we use an ADAM optimizer to update the network with an initial learning rate of 2e-4 and a batch size of 64.

\subsection{Results and Analysis}
In this section, we first discuss the contribution of $L_cls$ and $L_gan$ loss functions to joint semantic representation learning. Then we show the translated character trajectories and handwritten inertial signals, and analyze the quality of the translated data by feeding them into the pre-trained CNN model to check if they are identifiable. We also compare the performance of our model with CycleGAN using MMD (Maximum Mean Discrepancy) score\cite{xu2018empirical} and accuracy. Finally, we combine the translated data with the original data to improve classification performance.

\subsubsection{Illustration of Automatically Generated character trajectories and inertial signals}
\begin{figure}
\centering
\includegraphics[width=0.45\textwidth]{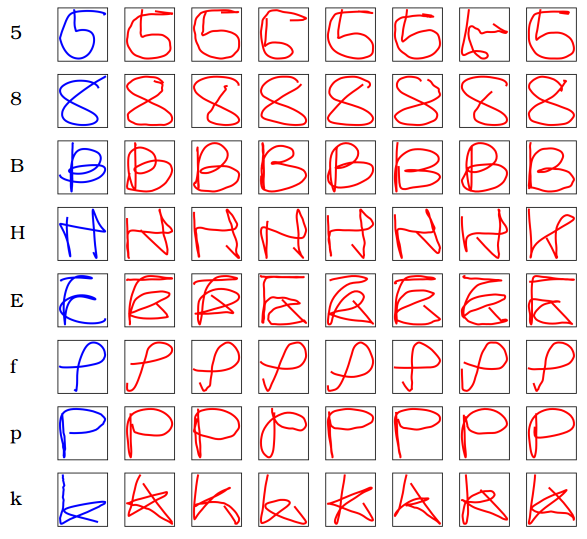}
\caption{Illustration of the translated characters for different classes under the first set of experimental settings ( Iner.-Tra.(6DMG)). Each row represents a particular character class. The blue trajectories are the paired ground truth of input inertial signal, the red trajectories are generated from the inertial signal corresponding to the ground truth on the left.}
\label{figure8}
\end{figure}

\begin{figure}[ht]
\centering
\subfigure{
\begin{minipage}[b]{0.45\textwidth}
\includegraphics[width=\textwidth]{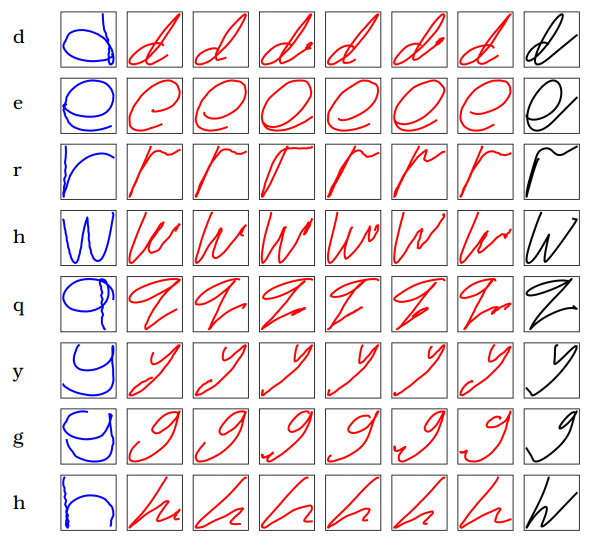}
\end{minipage}
}
\caption{Illustration of the translated characters under the second experimental settings. The blue trajectories are the paired ground truth of input inertial signal, the red trajectories are generated by Air-Writing Translater from the inertial data, the black ones are randomly selected from CT.}
\label{figure9}
\end{figure}

With the unsupervised domain adaptation, our Air-Writing Translater can either draw character trajectories or generate handwritten inertial signals. To verify the ability of drawing different character trajectories, Fig.\ref{figure8} and Fig.\ref{figure9} show the characters translated under the first and second sets of experimental settings, respectively. These figures demonstrate that our model succeeds in generating (drawing) meaningful and identifiable character trajectories while preserving the semantic content conveyed from the input inertial signal. As can be seen from Fig.\ref{figure8}, there is a high degree of similarity between the translated character and the input's paired ground truth. That is, if the samples of the source and target domains come from the same dataset handwritten in the same style, our model can achieve trajectory reconstruction. Thus, although trajectory collection is still required in the future, the strict synchronous collection of trajectories is not necessary anymore.

From Fig.\ref{figure9}, our network succeeds to transform 6DMG inertial signal with CT trajectories. Furthermore, the translated characters look much more like trajectories in CT than in 6DMG. As we expected, when the encoder compresses the inertial signal into latent space, it filters domain-dependent handwriting style content. In the meanwhile, the target decoder reconstructs it with the handwriting style in CT domain, acting like a handwriting style transfer. Ideally, we don't need to be involved in collecting trajectory at all, but instead can access the public handwritten trajectory dataset. We don't even need in-air handwriting trajectories, online handwritten trajectories of X-Y coordinates written on a trackboard/touchscreen are enough, and our model can transform inertial signals into trajectories under the target style.

These results verify not only the ability of the model in generating diverse and human-readable character trajectories but also the diversity of the model in handling different handwriting styles.

\begin{figure}
\centering
\subfigure{
\begin{minipage}[b]{0.45\textwidth}
\includegraphics[width=\textwidth]{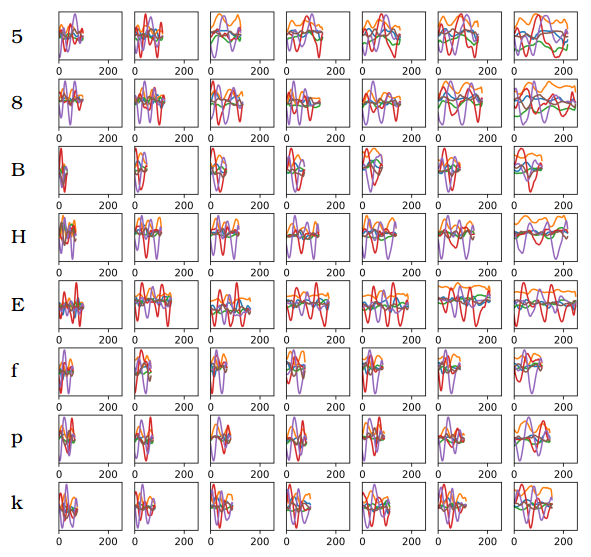}
\end{minipage}
}
\caption{Illustration of the translated Inertial samples for different classes under the third set of experimental settings (Tra.(6DMG)-Iner.). Each row represents a particular character class.}
\label{figure10}
\end{figure}

\begin{figure}[ht]
\centering
\subfigure{
\begin{minipage}[b]{0.48\textwidth}
\includegraphics[width=\textwidth]{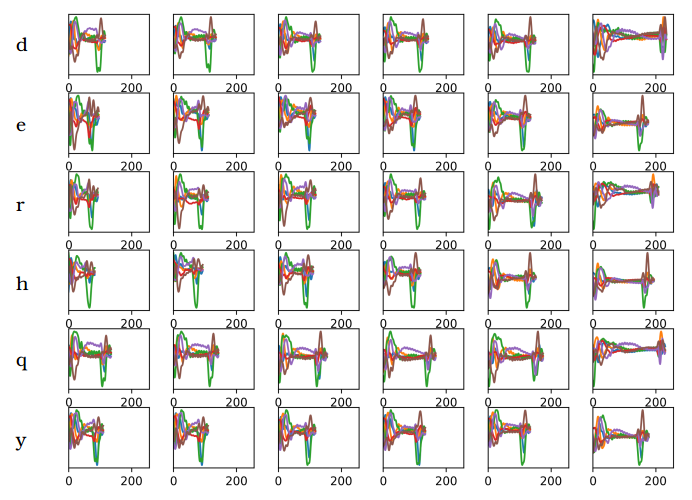}
\end{minipage}
}
\caption{Illustration of the translated Inertial samples under the fourth experimental settings.}
\label{figure11}
\end{figure}

To verify the ability of generating different handwritten inertial signals, Fig.\ref{figure10} and Fig.\ref{figure11} show the inertial signals translated from trajectories under the third and fourth sets of experimental settings, respectively. Fig.\ref{figure10} illustrates the samples of the inertial signals translated from the trajectory of 6DMG. Fig.\ref{figure11} illustrates some of the inertial signals translated from the trajectory of CT. As can be seen from these two figures, the translated inertial signals of the same character class have higher similarity but are not identical, while those of different classes have significant differences. These results show that our model learns the differences between different character classes while ensuring the diversity of the same class.

\subsubsection{Discussion on loss function}
\begin{figure}[ht]
\centering
\subfigure[]{
\begin{minipage}[b]{0.14\textwidth}
\includegraphics[width=\textwidth]{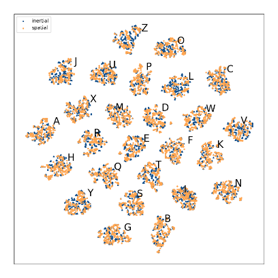}
\end{minipage}
}
\subfigure[]{
\begin{minipage}[b]{0.14\textwidth}
\includegraphics[width=\textwidth]{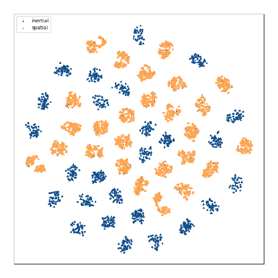}
\end{minipage}
}
\subfigure[]{
\begin{minipage}[b]{0.14\textwidth}
\includegraphics[width=\textwidth]{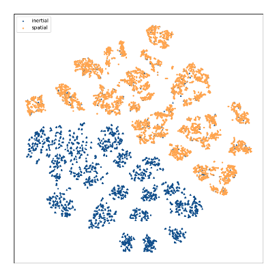}
\end{minipage}
}
\caption{t-SNE visualization of latent vectors. The blue points refer to inertial embeddings, and the yellow ones are trajectory embeddings. (a) trained with full losses, (b) trained with $L_{rec}$ and $L_{cls}$, without latent adversarial loss $L_{gan}$, (c) trained with $L_{rec}$ and $L_{gan}$, without latent classification loss $L_{cls}$}
\label{figure12}
\end{figure}

To verify the contribution of latent classification loss $L_{cls}$ and latent adversarial loss $L_{gan}$ to the joint semantic representation learning, we remove $L_{cls}$ and $L_{gan}$ respectively, and then use the remaining losses to train the network. We use t-SNE to reduce the latent dimension to 2 so that they can be visualized in a x-y coordinate. Fig.\ref{figure12}(a) shows the semantic embedding learned from a normal training network using $L_{gan}$ and $L_{cls}$. We can see that the samples of the same character class are clustered together, and different classes are far from each other. When we remove $L_{gan}$, as shown in Fig.\ref{figure12}(b), the embeddings are still distinguishable, but the semantic content of the two domains are not aligned, the network thus cannot perform the correct translation. Finally, Fig.\ref{figure12}(c) indicates that the adversarial training completely fails when we remove $L_{cls}$. We believe that the latent classification loss not only constrains on semantic consistency, but also guides and promotes a better adversarial training in latent space.

\subsubsection{Quality analysis}
We further analyze the quality of the translated samples in Inertia-to-Trajectory translation task. The pre-trained CNN model is used to check whether the translated trajectories are recognizable or not. We also use MMD score as an evaluation criterion to calculate the distance between the distribution of the translated trajectory and the paired ground truth to indicate the quality of the transfer results. In addition, we compare the quality of the trajectories translated by our model and CycleGAN model, given its reliable performance in unsupervised image-to-image translation. We have modified the architecture of CycleGAN to deal with in-air handwritten data. The generator consists of a down-sampling component with 3 convolutional layers, a feature extraction component with 6 residual blocks, and an up-sampling component with 3 deconvolutional layers. The discriminator includes 5 convolutional layers. We remove the fully connected structures so that it can accept input sequences of any length, and the kernels are strip-shaped like in Air-Writing Translater.

\begin{figure}
\centering
\includegraphics[width=0.48\textwidth]{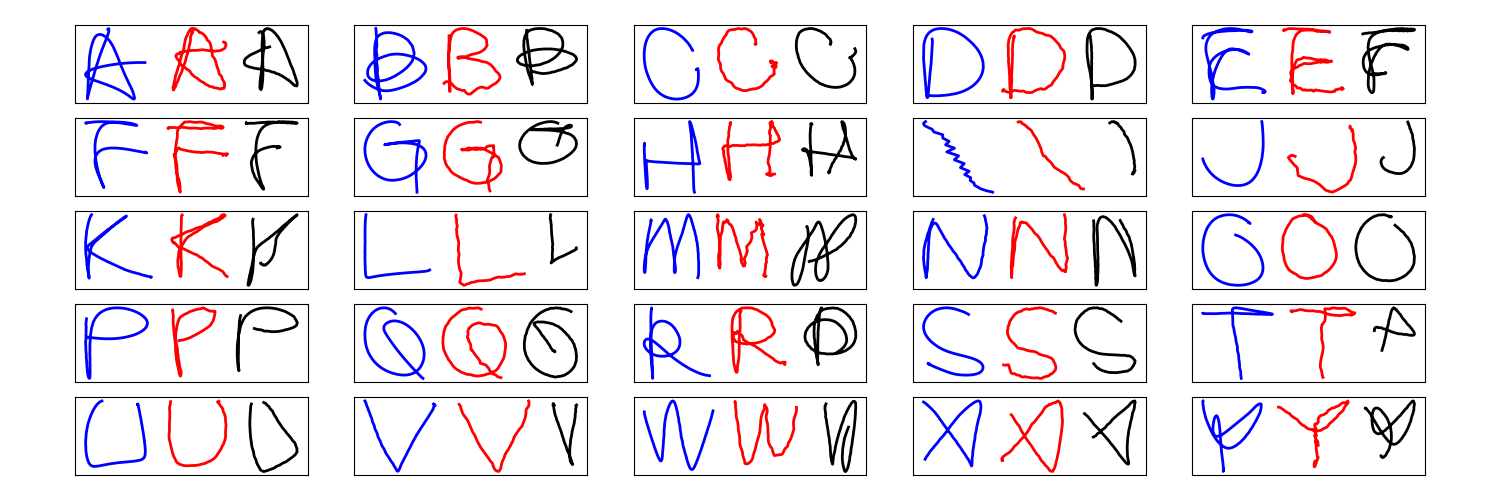}
\caption{Comparison between Air-Writing Translater and CycleGAN on inertia-to-trajectory translation task. The blue trajectories are the paired ground truth of input inertial signal, the red ones are outputs of Air-Writing Translater, the black ones are outputs of CycleGAN.}
\label{figure13}
\end{figure}

TABLE II shows the classification accuracies and losses for real samples, samples translated by Air-Writing translater, and samples translated by CycleGAN on the 62 classes. TABLE III gives a comparison of MMD score. As revealed in TABLE II, the accuracy of trajectories translated by our model can match the accuracy of real samples. This verifies the ability of our model to correctly and successfully translate inertial samples to trajectory samples. From TABLE II and III, compared with CycleGAN, our model has better classification accuracy and MMD score. Fig.\ref{figure13} illustrates some trajectories translated by our model and CycleGAN. According to the trajectories translated by CycleGAN (the black one in Fig.\ref{figure13}), it seems feasible to conduct a sample-level adversarial training. But the lack of semantic guidance is the reason why CyclGAN's reconstructed trajectories are not good. After all, there is almost no shared pixel structure between the inertial signal and motion trajectory. The CycleGAN may achieve better performance, at the cost of further tuning of the architecture and hyper-parameters.

\begin{table}
\caption{Comparison of the classification accuracy and loss for real samples, samples translated by our model, and samples translated by CycleGAN.}
\centering
\begin{tabular}{cccc}
\hline
 & real samples & \tabincell{c}{samples translated\\by our model} &  \tabincell{c}{samples translated\\by cycleGAN}\\
\hline
accuracy& 0.988& 0.9659& 0.6657\\
loss& 0.1354& 0.2938& 2.56\\
\hline
\end{tabular}
\end{table}

\begin{table}
\caption{Comparison of MMD scores}
\centering
\begin{tabular}{ccc}
\hline
 & \tabincell{c}{samples translated\\by our model}& \tabincell{c}{samples translated\\by cycleGAN}\\
\hline
MMD score& 0.0277& 0.1011\\
\hline
\end{tabular}
\end{table}

\subsubsection{A further step toward in-air handwriting recognition}
To further evaluate the translated samples, we design a two-stream ConvNet model that combines the original real samples in one domain with the translated samples in the other domain to see if it improves the performance of in-air handwriting recognition. For comparison, we also trained a CNN classification model for real samples. We conduct experiments on 6DMG and CT datasets, respectively. TABLE IV and V report the average accuracy of each dataset and the accuracy of each class.

For the 62 character classes in 6DMG, adding translated samples can effectively improve the performance of in-air handwriting recognition. Especially for the trajectory samples, when combined with the translated inertial samples, the accuracy is significantly improved by 4.69\%, as shown in Table IV. As described in "Datasets" section, the sample distribution in 6DMG dataset is not balanced. The number of samples in lowercase letters and Arabic numerals is much smaller than that in uppercase letters. Fig.\ref{figure14}(a) and (c) indicate that the samples generated by our domain adaptation model contribute significantly to the recognition performance of minority classes. In general, in an unbalanced dataset, the classification model tends to guarantee the accuracy of majority classes, especially when the distance between certain categories is too close in the feature space. For example, 0 and o are often misclassified into O. But, by combing information from another domain, the distance between some confused classes can be pulled away in a new feature space. Besides, Fig.\ref{figure14}(c) achieves a better improvement than Fig.\ref{figure14}(a), which means inertial data plays a more important role than trajectory data in in-air handwriting recognition. This also means that, compared with the cost of collecting labeled inertial samples, we can get a classifier with better performance at a much lower cost.

For the 20 character classes in CT, adding translated samples can also improve recognition performance, even if the data of the two domains for translation task come from different datasets. Since the baseline accuracy is already high, the performance improvement does not look as significant as in the 62 classes. Fig.\ref{figure14}(b) and (d) give the comparison of the accuracies of each class for CNN and two-stream ConvNet. In Fig.\ref{figure14}(b), for real inertial data, the characters "n" and "o" are misclassified as "h" and "a", respectively. Because there is no visual feedback of the written trajectory during in-air handwriting, these characters with similar motion paths (such as "n" and "h", "o" and "a") are often confused. Therefore, by adding translated trajectory samples, these confusing characters become distinguishable. In Fig.\ref{figure14}(d), for real trajectory data, misclassification occurs in characters "h", "n", "w", "u", and "m" because they are similar in the handwirtten trajectory (see Fig.\ref{figure14}(b)). After addting the translated inertial samples to distinguish, the recognition performance is improved.

\begin{table}
\caption{Comparison of classification performance for 62 classes in 6DMG}
\centering
\begin{tabular}{cccc}
\hline
Translation & \tabincell{c}{Real sample\\+CNN}& \tabincell{c}{Real sample\\+Translated sample\\+two stream ConvNet}& \tabincell{c}{accuracy on\\each class}\\
\hline
Iner.-to-Tra. & 0.9653& 0.9699& \tabincell{c}Fig.\ref{figure14}(a)\\
\hline
Tra.-to-Iner. & 0.9190& 0.9659& \tabincell{c}Fig.\ref{figure14}(c)\\
\hline
\end{tabular}
\end{table}

\begin{table}
\caption{Comparison of classification performance for 20 classes in CT}
\centering
\begin{tabular}{cccc}
\hline
Translation & \tabincell{c}{Real sample\\+CNN} & \tabincell{c}{Real sample\\+Translated sample\\+two stream ConvNet}& \tabincell{c}{accuracy on\\each class}\\
\hline
Iner.-to-Tra. & 0.9922& 1.0000& Fig.\ref{figure14}(b)\\
\hline
Tra.-to-Iner. & 0.9939& 1.0000& Fig.\ref{figure14}(d)\\
\hline
\end{tabular}



\footnotesize{$^a$ Inertial data comes from 6DMG and trajectory data comes from CT.}
\end{table}

\textbf{\begin{figure}[ht]
\centering
\subfigure[]{
\begin{minipage}[]{0.49\textwidth}
\includegraphics[width=\textwidth]{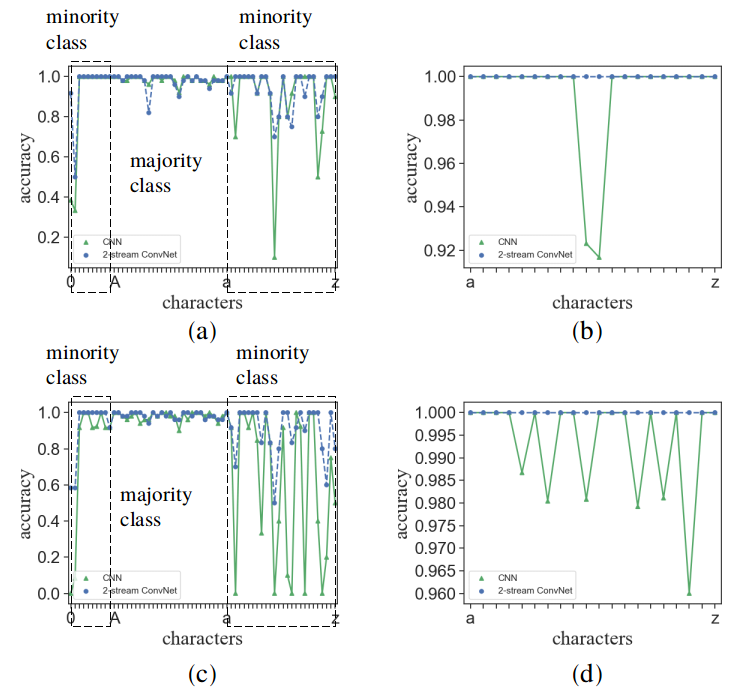}
\end{minipage}
}
\caption{(a)(b)(c)(d) corresponds to the results stated in TABLE III and TABLE IV.}
\label{figure14}
\end{figure}}

\section{Conclusions}
Inertia-Trajectory translation of in-air handwriting is a very challenging problem that has not been thoroughly studied in the literature. In this work, we propose a novel domain adaptation model named Air-Writing Translater to solve unsupervised Inertia-Trajectory translation problem. The model constructs a latent space to map inertial and trajectory samples into semantic representations. By introducing feature-level adversarial training, the model learns to translate between inertial data and handwritten trajectory with its semantic content preserved. The network architecture is carefully designed to handle time series, so that it can accept input sequence of arbitrary length, and support translation between domains at different sampling rates. Experiments on two public datasets show that our model performs reliable and effective cross-domain translation. The results show that the translated samples can be identified by the CNN model with high accuracy. To our best knowledge, Air-Writing Translater is the first domain adaptation method that achieves outstanding performances in unsupervised Inertia-Trajectory translation. Other than translating between inertial data and trajectory, an interesting future direction is to extend the proposed model to convert between static images and dynamic inertial signals, which is a hard problem and has great value in practical applications.

\bibliographystyle{IEEEtran}
\bibliography{reference}
\ifCLASSOPTIONcaptionsoff
  \newpage
\fi

\begin{IEEEbiography}{Songbin Xu}
Biography text here.
\end{IEEEbiography}

\begin{IEEEbiographynophoto}{Yang Xue}
Biography text here.
\end{IEEEbiographynophoto}


\begin{IEEEbiographynophoto}{Xin Zhang}
Biography text here.
\end{IEEEbiographynophoto}




\end{document}